\documentclass[journal]{IEEEtran}
\usepackage{cite}
\usepackage{amsmath}
\usepackage{booktabs}
\usepackage{graphicx}
\usepackage{multirow}
\usepackage{placeins} 
\usepackage{float}
\usepackage{stfloats}
\usepackage{xcolor}
\usepackage{algorithm}
\usepackage{algorithmic}
\usepackage{array}
\usepackage{cleveref}
\usepackage{amsmath}
\usepackage{amssymb}
\usepackage{newtxtext,newtxmath}

\ifCLASSINFOpdf
\else
\fi
\begin{document}
%
\title{STS-NET: Spatio-Temporal Stress Network for Self-Supervised Crop Stress Detection using Satellite Image Time Series}
%
%


\author{Pradeep Dalal,  Rajiv Ranjan, Sushil Ghildiyal, 
        Shashank Tamaskar, and Neeraj Goel,~\IEEEmembership{Member,~IEEE}
        
\thanks{P. Dalal, S. Ghildiyal, and N. Goel are with the Department of Computer Science and Engineering, Indian Institute of Technology
 Ropar, Rupnagar 140001, India (e-mail: pradeep.dalal283@gmail.com;
 sushil.20csz0021@iitrpr.ac.in; neeraj@iitrpr.ac.in)}
\thanks{R. Ranjan and S. Tamaskar are with the Department of Robotics and
 Autonomous Systems, Plaksha University, Mohali 140306, India (e-mail:
 rajiv.ranjan@plaksha.edu.in; shashank.tamaskar@plaksha.edu.in).}
}

\maketitle

\begin{abstract}
Early and accurate detection of crop stress is essential to improve agricultural productivity and ensure global food security. However, collecting a large labeled crop stress dataset is a challenging task. To address this challenge, we introduce a novel spatial-temporal stress network (STS-NET), built on a self-supervised 3D-convolutional autoencoder (3D-CAE), designed to utilize Satellite Image Time Series (SITS) data for crop stress detection. STS-NET exploits four vegetation indices: Normalized Difference Vegetation Index (NDVI), Normalized Difference Vegetation Index (GNDVI), Red-Edge Chlorophyll Index (RECI) and Normalized Difference Red-Edge Index (NDRE) obtained from high resolution Planetscope imagery to capture spatiotemporal stress patterns. The model is trained on our BSPT (Barnala Spatial-Temporal) dataset and evaluated on a real-world sugarcane dataset collected over a year from a 2.5-acre test plot located in Lakhimpur-Kheri (LK) district in Uttar Pradesh in India. STS-NET achieved a precision of 97. 98\% for water stress, 85.08\% for nitrogen stress, and 83.47\% for combined stress. The results demonstrate the potential
of STS-NET in effectively detecting stress in sugarcane crops with minimal reliance on labeled data. Furthermore, STS-NET can serve as a robust feature extractor for simpler models.


\end{abstract}


%
\IEEEpeerreviewmaketitle

\section{Introduction}

 \IEEEPARstart{W}{ith} the gradual rise in global population, the demand for food is steadily increasing, while cultivable land
remains limited. Hence, improving the productivity of existing farmlands is essential to ensure food security. Among various limiting factors, stress due to water and nitrogen deficiencies is a major contributor to yield loss. Early detection of such stress is critical, as delayed intervention can significantly reduce both yield and quality. However, conventional field based monitoring is labor intensive and time consuming and
lacks spatial scalability.

Remote sensing offers a scalable solution, with Vegetation Indices (VI) derived from multispectral or hyperspectral im
agery effectively capture crop physiological changes. Indices such as NDVI, GNDVI, NDRE, and RECI have been widely used to estimate biomass, chlorophyll, and detect water and
nitrogen stress \cite{barboza2023performance, burns2022determining, boiarskii2022application}.  In parallel, deep learning methods have shown a strong potential for crop stress modeling by learning
complex spatio-temporal patterns directly from data. When combined with high-resolution remote sensing imagery, deep learning models can enable fine-grained detection of stress
at scale. Nampally et al. \cite{nampally2023stressnet}  used multi-spectral UAV imagery with CNN and BiLSTM-attention models to assess water stress
in maize. Similarly, Zhang et al. \cite{zhang2019mapping} used Sentinel-2 time series and a GRU-based network to detect heavy metal-induced stress in rice. In this study, we focus on sugarcane, a commercial crop used for sugar, ethanol, and by-products \cite{dotaniya2016use}. It is a long
duration, water-intensive and nutrient-intensive crop that is highly sensitive to environmental factors. Any imbalance in
these inputs can lead to significant reductions in both crop quality and yield \cite{mehdi2024factors,10640589}. Due to its high agronomic value and vulnerability to abiotic stresses such as water and nitrogen
stress, sugarcane requires continuous monitoring to maintain yield and quality. Most existing studies primarily focus on labeled datasets and use low spatial resolution satellite imagery such as Sentinel-2 \cite{lehouel2024remote} to assess crop water stress. Moreover, the high temporal resolution in these works is often limited, using only a small number of time-series images for a given
region over time. Furthermore, previous research has focused on the detection of water stress \cite{zhang2019mapping} or nitrogen stress \cite{elmetwalli2020estimation} individually, with limited efforts directed toward understanding the combined impact of both stressors on crop health. The key
contributions of our work are as follows:

\begin{itemize}
\item We propose STS-NET, a novel self-supervised learning framework based on a 3D Convolutional Autoencoder (3D-CAE) for extracting spatio-temporal features from unlabeled satellite image time series.
\item We introduce the BSPT dataset, a large-scale unlabeled SITS dataset derived from PlanetScope imagery, used for pretraining the 3D-CAE model.
\item We validate our approach on a real-world experimental dataset from a controlled sugarcane farm in Lakhimpur Kheri (Uttar Pradesh, India), where stress conditions were systematically induced.
\end{itemize}

    

\section{Literature Review}

Efficient detection of crop stress is critical to improving agricultural productivity. The need for timely crop health and water management has been highlighted in \cite{dutta2020improved}, with particular attention to water stress due to its role in optimizing irrigation \cite{safdar2023crop}. In addition to water, nitrogen is an essential nutrient for sugarcane growth and sugar accumulation during the maturity stage, and is traditionally measured by labor intensive leaf sampling and laboratory analysis \cite{abdel2008imaging}. Similarly, conventional water stress detection methods involve ground-based measurements, which are time-consuming and spatially constrained \cite{katimbo2022crop}. Although thresholding-based techniques exist, they are sensitive to crop type and environmental variation \cite{fu2022critical}. 

Vegetation indices derived from multispectral satellite imagery have shown promise in estimating crop physiological stress. Studies have identified GNDVI and RENDVI as sensitive to nitrogen stress in crops such as maize and tea \cite{samarina2024efficient, ramachandiran2017relationship}, while NDWI has been effective in detecting water stress for wheat \cite{parmar2020wheat}.

Recent advances in deep learning have enabled automated stress detection using satellite data. Random forest and SVM models have been used to estimate the leaf nitrogen content from Sentinel-2 imagery \cite{soltanikazemi2022field}, and multi-temporal imagery have been used to detect heavy metal-induced stress in rice \cite{liu2018heavy}. Deep learning models such as DenseResUnet have been applied to water stress detection from single-timestep images \cite{virnodkar2021denseresunet}, and CNN-ViT hybrids have been used with multi-temporal UAV and Landsat-8 thermal data \cite{lehouel2024remote}. Recent works like Presto \cite{tseng2023lightweight} introduce a self-supervised transformer-based approach that can be used as a feature extractor for various remote sensing tasks. Although Singh et al. \cite{singh20243d} employed a 3D-CAE for unsupervised crop type mapping, its application to crop stress detection, particularly in a self-supervised setting, remains unexplored. Most researchers used low-resolution labeled datasets with limited temporal coverage, usually addressing nitrogen or water stress separately, neglecting their combined effects on crop health \cite{soltanikazemi2022field,virnodkar2021denseresunet,lehouel2024remote}. Our method removes the reliance on labeled data by employing self-supervised learning and applies a 3D-CAE to capture both spatial and temporal dynamics from high-resolution satellite image time series. To our knowledge, this is the first approach to explore self-supervised crop stress detection using 3D-CAE and SITS data.

\section{Dataset Description}
\subsection{Lakhimpur Kheri Dataset}
The Lakhimpur Kheri (LK) dataset, developed by Ranjan et al. \cite{10.1007/978-3-031-88217-3_30, 10984153, shikhar2024evaluation}, originates from a controlled field experiment conducted in Lakhimpur Kheri, Uttar Pradesh (28°27'14.2"N, 80°55'42.0"E), covering approximately 3 acres. The dataset consists of 62 individual sugarcane fields,
as shown in Fig 1, each of which was subjected to specific combinations of water and nitrogen stress as part of a planned
experimental setup. The controlled experiment was performed on Plant Cane(first-year sugarcane) planted in February 2023.
All water and nitrogen treatments were imposed during the tillering and grand-growth stages (April–September), which are
physiologically most sensitive to stress. No ratoon cropping was part of the experiment. The study area was divided into three
primary blocks based on water availability : i) Block 1 (B1): High water block, ii) Block 2 (B2): Medium water block, iii) Block 3 (B3): Low water block. Within each of these blocks, nitrogen stress was induced through controlled fertilizer application,
divided into three levels: i) P: Low Nitrogen block, ii) Y: Medium Nitrogen block, iii) R: High Nitrogen block. Nitrogen stress levels were induced by applying varying amounts of fertilizer: 0.25kg, 0.5kg, and 1kg for blocks P, Y, and R in Block 1, and 0.2kg, 0.4kg, and 0.8kg for the corresponding blocks in Blocks 2 and 3. The Water levels were adjusted relative to the standard irrigation amount, with 150\% for high water treatment, 100\% for medium, and 50\% for low water blocks. These experimental conditions resulted in nine distinct classes, each representing a unique combination of nitrogen and water stress.


\begin{figure}
    \centering
    \includegraphics[width=1\linewidth]{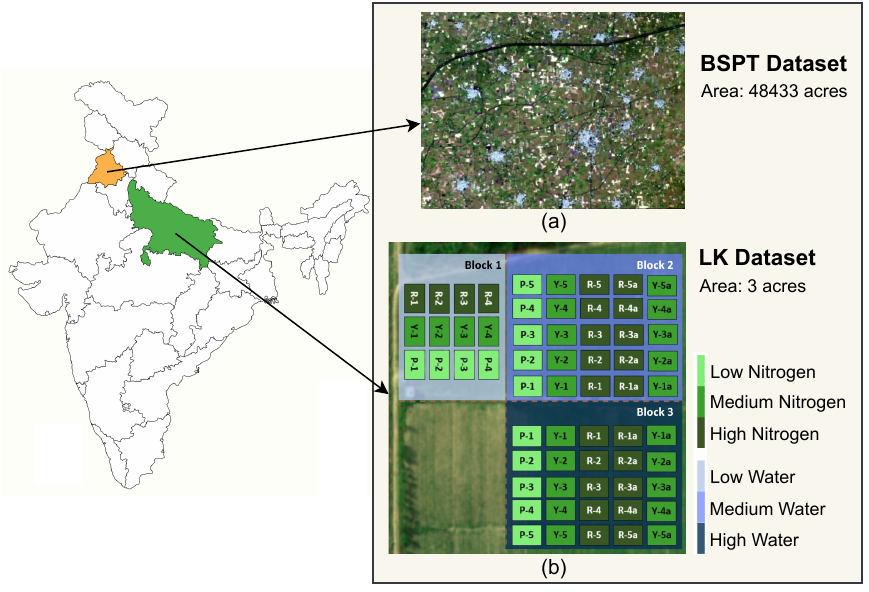}
    \caption{(a) BSPT dataset (Punjab, India) collected from PlanetScope imagery, and (b) Test farm layout of the LK (Lakhimpur Kheri, UP, India) dataset \cite{10.1007/978-3-031-88217-3_30, 10984153, shikhar2024evaluation}.}    
    
    \label{fig:study_area}
\end{figure}

\subsection{BSPT Dataset}
The dataset for pretraining was gathered using satellite images obtained from the Barnala district in Punjab, located
at approximately 30°35’56.94”N, 75°27’34.35”E. The overall
area designated for this purpose covers around 196 km\textsuperscript{2}, featuring a variety of agricultural land with differing cropping patterns and environmental factors. The satellite imagery
was collected from the PlanetScope satellite constellation, which is operated by Planet Labs. A total of 32 time-series images were taken from February 2023 to February 2024 
captured at ~9–11 days interval, each with less than 15\% cloud cover. The dataset consists of approximately 22 million pixels, representing a broad and diverse sample of
cultivated land across multiple seasons and growth stages.


\begin{figure*}
    \centering
    \includegraphics[width=\textwidth]{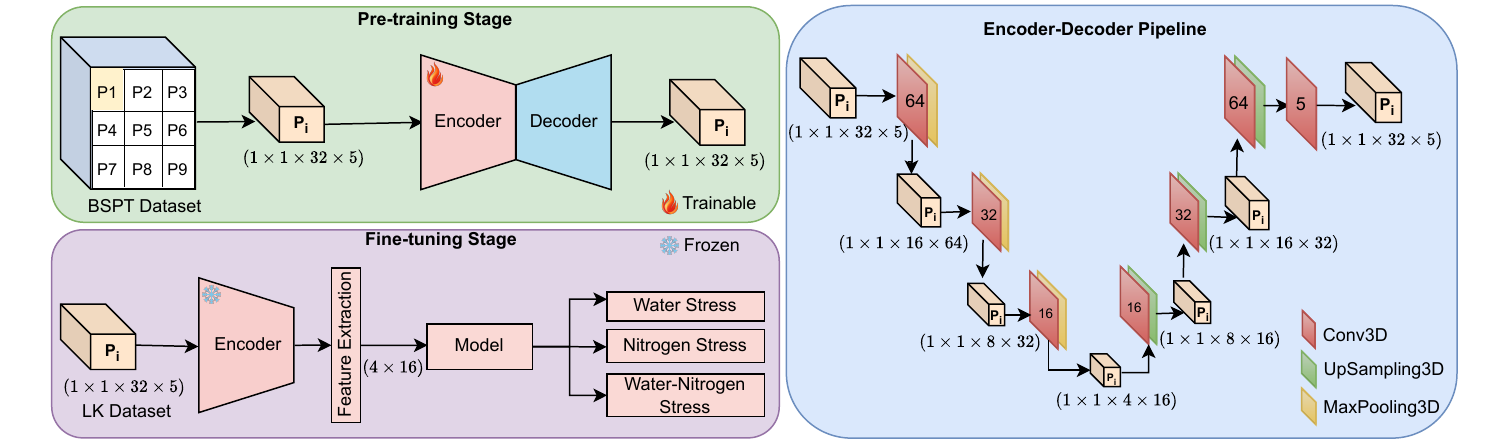}
    \caption{Represents the STS-NET architecture pipeline for stress detection in sugarcane farms. In pretraining stage, the autoencoder was trained on unlabelled data, and in the Finetuning stage, the trained encoder was utilized to extract the best feature map from the labelled data. Later, different models were used to classify water and nitrogen stress.}
    \label{fig:stsnet_pipeline}
\end{figure*}

\section{Methodology}
The entire methodology is divided into multiple stages. Initially, satellite image time series data was preprocessed by using python libraries. In the next stage, a 3D-CAE was then pretrained using a self-supervised pixel reconstruction task to extract rich spatio-temporal features. Later, a classification head is attached to the pretrained model for classifying the pixels into their particular stress-related class.

\subsection{Data Preprocessing}
The preprocessing pipeline began by mosaicking multiple satellite image tiles into unified composite images for each timestamp using Rasterio, resulting in 32 temporally sequenced .tif files. Missing pixel values were imputed using a mean-based strategy. Vegetation indices (NDVI, GNDVI, RECI, NDRE) were computed from spectral bands and selected based on their physiological relevance and superior performance in ablation studies (Table II). Specifically, NDVI reflects canopy vigor and water availability \cite{safdar2023crop}, GNDVI correlates with chlorophyll and nitrogen status \cite{burns2022determining}, RECI captures red-edge chlorophyll sensitivity, and NDRE is sensitive to early chlorophyll loss from both water and nitrogen stress \cite{boiarskii2022application}. To address class imbalance, stratified sampling was applied during train–validation splitting. The final dataset was organized as spatio-temporal pixel-level time series vectors for model training.
\subsection{Self-Supervised Pretraining with 3D Convolutional Autoencoder}

The dataset used for pretraining in this study is the BSPT dataset. To extract meaningful patterns from the large volume of unlabeled satellite data, we trained a 3D-CAE in a self-supervised manner. The 3D-CAE architecture consists of an encoder ($f_\theta$) and a decoder ($g_\phi$). The aim of this pretraining stage (as illustrated in Fig \ref{fig:stsnet_pipeline}) is to learn a feature extractor $f_\theta$, that captures both the spatial and temporal dynamics of crop development and stress progression without relying on any manual labelled data. The objective is to reconstruct the original sequence of a pixel’s spectral values by using information from the surrounding space and time. The input to the autoencoder is a single pixel tracked across time, represented as a tensor:
\begin{align}
x &= \left\{ \mathbf{t}_i \;\middle|\; i = 1, \dots, 32 \right\}, \notag \\
\mathbf{t}_i &= \left[ t_i^{R},\ t_i^{G},\ t_i^{B},\ t_i^{\text{NIR}},\ t_i^{\text{VIs}} \right], \quad
x \in \mathbb{R}^{1 \times 1 \times T \times C}
\end{align}
where $T = 32$ denotes the number of timesteps, and $C = 5$ denotes the number of input channels. Here $t_i^{\text{VIs}} = \frac{1}{4} \left( \text{NDVI}_i + \text{GNDVI}_i + \text{RECI}_i + \text{NDRE}_i \right)$ denotes the average vegetation index at timestep $i$. The encoder of the 3D-CAE consists of stacked 3D convolutional layers that gradually compress this input pixel-timeseries into a lower-dimensional latent feature space. These layers capture temporal dependencies and inter-band relationships across the input sequence, and the decoder is used to reconstruct the original pixel timeseries sequence.


The model is optimized using the Mean Squared Error (MSE) loss. MSE can be represented by:
\[
\textstyle
MSE = \frac{1}{N \times T \times C} \sum_{n=1}^{N} \sum_{t=1}^{T} \sum_{c=1}^{C} \left( x^{(n)}_{t,c} - \hat{x}^{(n)}_{t,c} \right)^2
\]

where, \( x^{(n)}_{t,c} \) is the true spectral value of the pixel \( n \) at time \( t \) and channel \( c \),  \( \hat{x}^{(n)}_{t,c} \) is the corresponding reconstructed value predicted by the model, and N represents the total number of pixels present in a single training batch. As the model stabilises, a feature vector is created by encoding each temporal stack of pixels used for classification purposes.

\begin{algorithm}[H]
\caption{STS-NET: Self-Supervised Stress Detection Framework}
\label{alg:stsnet}
\begin{algorithmic}[1]
\REQUIRE Unlabeled time-series data $\{\mathbf{x}_n\}_{n=1}^{N}$, Labeled data $\{(\mathbf{x}_m, y_m)\}_{m=1}^{M}$, where $\mathbf{x} \in \mathbb{R}^{T \times C}$

\STATE \textbf{Pretraining Stage (Self-Supervised)}
\STATE Initialize 3D-CAE with encoder $f_\theta$ and decoder $g_\phi$
\FOR{each $\mathbf{x}_n$ in unlabeled data}
    \STATE $\hat{\mathbf{x}}_n \gets g_\phi(f_\theta(\mathbf{x}_n))$
    \STATE Compute loss: $\mathcal{L}_{\text{MSE}}(\mathbf{x}_n, \hat{\mathbf{x}}_n)$
\ENDFOR
\STATE Update $f_\theta$, $g_\phi$ to minimize $\mathcal{L}_{\text{MSE}}$

\STATE \textbf{Fine-tuning Stage (Supervised)}
\STATE Discard $g_\phi$, freeze $f_\theta$
\FOR{each $\mathbf{x}_m$ in labeled data}
    \STATE $\mathbf{z}_m \gets f_\theta(\mathbf{x}_m)$
\ENDFOR
\STATE Train classifier $h(\cdot)$ (e.g., RF, XGBoost) on $\{\mathbf{z}_m, y_m\}$

\end{algorithmic}
\end{algorithm}

\begin{table*}[t]  
    \centering
    \caption{Performance comparison of models across Water, Nitrogen, and Water-Nitrogen stress}
    \label{tab:grouped_metrics}
    \resizebox{\textwidth}{!}{%
    \begin{tabular}{lcccccccccccc}
        \toprule
        \multirow{2}{*}{\textbf{Model Name}} & \multicolumn{4}{c}{\textbf{Water Stress}} & \multicolumn{4}{c}{\textbf{Nitrogen Stress}} & \multicolumn{4}{c}{\textbf{Water-Nitrogen Stress}} \\
        \cmidrule(lr){2-5} \cmidrule(lr){6-9} \cmidrule(lr){10-13}
         & Accuracy & Precision & Recall & F1-Score & Accuracy & Precision & Recall & F1-Score & Accuracy & Precision & Recall & F1-Score \\
        \midrule
        Random Forest (Supervised) & 0.59 & 0.57 & 0.59 & 0.57 & 0.54 & 0.52 & 0.54 & 0.53 & 0.31 & 0.28 & 0.31 & 0.29 \\
        SVM (Supervised) & 0.44 & 0.53 & 0.44 & 0.46 & 0.38 & 0.51 & 0.38 & 0.39 & 0.23 & 0.25 & 0.23 & 0.20 \\
        XGBOOST (Supervised) & 0.60 & 0.59 & 0.60 & 0.59 & 0.54 & 0.53 & 0.54 & 0.53 & 0.28 & 0.26 & 0.28 & 0.26 \\
        CNN (Supervised) & 0.54 & 0.56 & 0.54 & 0.51 & 0.52 & 0.41 & 0.52 & 0.46 & 0.28 & 0.26 & 0.28 & 0.25 \\
        Presto \cite{tseng2023lightweight} (Foundational Model) & 0.82 & 0.79 & 0.78 & 0.78 & 0.81 & 0.78 & 0.76 & 0.77 & 0.72 & 0.70 & 0.68 & 0.69 \\
        \textbf{3D-CAE + SVM (Proposed)} & 0.93 & 0.91 & 0.94 & 0.92 & 0.71 & 0.71 & 0.73 & 0.72 & 0.70 & 0.71 & 0.70 & 0.71 \\
        \textbf{3D-CAE + RF (Proposed)} & \textbf{0.97} & 0.98 & 0.98 & 0.98 & 0.85 & 0.85 & 0.85 & 0.85 & 0.83 & 0.84 & 0.83 & 0.83 \\
        \textbf{3D-CAE + XGBOOST (Proposed)} & 0.96 & 0.95 & 0.95 & 0.95 & \textbf{0.87} & 0.87 & 0.87 & 0.87 & \textbf{0.84} & 0.85 & 0.85 & 0.84 \\
        \bottomrule
    \end{tabular}
    }
\end{table*}

\subsection{Supervised Fine-Tuning}
After pretraining, the decoder
$g_\phi$ is discarded, and the 3D-CAE encoder is retained as a fixed feature extractor. Each labeled pixel from the dataset in \cite{shikhar2024evaluation} comprising 62 sugarcane plots from a controlled experiment in Lakhimpur Kheri, Uttar Pradesh is encoded into a 64-dimensional latent vector capturing its spatial-temporal characteristics. These feature vectors are used to train lightweight classifiers: Random Forest and XGBoost. In this study, water, nitrogen, and combined water-nitrogen stress are modeled as separate multiclass classification tasks. Water stress is defined across three irrigation levels: High (B1), Medium (B2), and Low (B3); nitrogen stress follows a similar 3-class categorization: High (R), Medium (Y), and Low (P). The combined stress task includes all nine permutations of water–nitrogen regimes, resulting in a 9-class label. Accordingly, each pixel is annotated with three labels water, nitrogen, and combined stress and separate classifiers are trained for each using the same encoder-derived features. Final predictions are validated using ground-truth samples from the experimental field.

\section{Results and Discussions}

\subsection{Experimental Setup}
The 3D-CAE model was implemented in Python 3 using TensorFlow and trained on an NVIDIA DGX A100 GPU (40GB). PlanetScope imagery was preprocessed with min–max normalization, and raster processing and vegetation index computation were performed using Rasterio, GeoPandas, and NumPy. The dataset was split 80:20 for training and testing, with stratified sampling applied to address class imbalance. Through extensive experimentation, the optimal configuration was found to be a batch size of 256, a learning rate of 0.001, and the Adam optimizer. The model was trained for up to 40 epochs with early stopping based on validation performance. 
\subsection{Experimental Results}

\subsubsection{Supervised Learning Using Lakhimpur Kheri Dataset}

All experiments in this section were conducted on pixel-level time-series data by utilizing the Lakhimpur Kheri dataset. We began by applying traditional supervised machine learning algorithms such as Random Forest and Support Vector Machine (SVM). These algorithms performed well, for higher-dimensional data, but they failed to capture the temporal relationship present in time series data. Next, we explored a supervised deep learning approach using a two layered CNN architecture, but due to the limited number of labeled samples, the CNN struggled to generalize well.\\

\subsubsection{Self-Supervised Presto Fine-Tuned on Lakhimpur Kheri Dataset}

To overcome the issue of scarce labeled data and further evaluate the effectiveness of self-supervised learning, we used Presto \cite{tseng2023lightweight}, a self-supervised model trained on pixel-level time series data. Presto outperformed conventional models as well as supervised CNN, demonstrating the crucial role of self-supervised learning in such tasks. We operated Presto in three different settings, where the features extracted from the model were passed to three classifiers: Random Forest, SVM, and XGBOOST for experimentation. Among these, the Presto + Random Forest configuration yielded the best performance, and the corresponding results are reported in Table \ref{tab:grouped_metrics}. Since Presto is trained on global landcovers pixels, it's performance can be improved by Agri-specific training.\\

\subsubsection{Self-Supervised 3D-CAE with BSPT and Lakhimpur Kheri Datasets}

To enhance spatio-temporal feature learning, we pretrained our model on the BSPT dataset and fine-tuned it on the Lakhimpur Kheri dataset. The proposed STS-NET framework integrates a self-supervised 3D Convolutional Autoencoder (3D-CAE) for feature extraction with conventional classifiers, outperforming all baselines. Among the model variants, 3D-CAE + Random Forest yielded the best performance for water stress classification, while 3D-CAE + XGBOOST was superior for nitrogen and combined water-nitrogen stress. These results demonstrate the 3D-CAE's ability to extract deep spatio-temporal representations from satellite data, significantly enhancing traditional classifiers over raw input-based models. Remarkably, our architecture operates with only 19.2K parameters substantially lower than Presto's 400K \cite{tseng2023lightweight} highlighting its efficiency in both accuracy and computational cost. Performance gains are attributed to the use of higher spatial resolution imagery and denser temporal sampling, enabling better characterization of stress dynamics in sugarcane. Ablation studies (Table~\ref{tab:ablation_study}) evaluated the impact of temporal depth and vegetation indices, confirming that combining RGB, NIR, and the mean of four VIs (NDVI, NDRE, RECI, GNDVI) yields optimal results. The model maintains robust performance with just 8 time steps (~2 months), indicating effective learning even with limited temporal input. While developed for sugarcane, the model’s resilience at lower temporal depths suggests applicability to shorter-duration crops, with accuracy improving further as more temporal snapshots are incorporated.

\begin{table}
\centering
\caption{Ablation Study on Vegetation Indices and Temporal Depth (T) for Stress Classification. \textit{*Avg(VIs) refers to the average of NDVI, NDRE, RECI, and GNDVI vegetation indices.}}
\label{tab:ablation_study}
\renewcommand{\arraystretch}{1.3}
\scriptsize
\begin{tabular}{|p{2.8cm}|c|ccc|c|}
\hline
\textbf{Input Config} & \textbf{T} & \multicolumn{3}{c|}{\textbf{Accuracy (\%)}} & \textbf{Channels} \\
\cline{3-5}
& & \textbf{Water} & \textbf{Nitrogen} & \textbf{WNS} & \\
\hline
RGB + NIR & 32 & 86.32 & 72.14 & 68.21 & 4\\
RGB + NDVI & 32 & 89.47 & 74.57 & 71.03 & 4\\
RGB + NIR + NDVI & 32 & 91.48 & 78.76 & 75.37 & 5 \\
RGB + NIR + Avg(VIs)* & 32 & \textbf{97.98} & \textbf{85.08} & \textbf{83.47} & 5\\
\hline
RGB + NIR + Avg(VIs)* & 8 & 90.32 & 70.97 & 65.73 & 5\\
RGB + NIR + Avg(VIs)* & 16 & 90.73 & 77.82 & 74.60 & 5\\
RGB + NIR + Avg(VIs)* & 24 & 93.55 & 82.26 & 81.45 & 5\\
RGB + NIR + Avg(VIs)* & 32 & \textbf{97.98} & \textbf{85.08} & \textbf{83.47} & 5\\
\hline
\end{tabular}
\end{table}


\section{Conclusion and Future Work}
This study presents STS-NET, a self-supervised stress detection framework leveraging a 3D Convolutional Autoencoder (3D-CAE) trained on Satellite Image Time Series (SITS) data. By capturing rich spatiotemporal representations from vegetation indices, the model effectively detects crop stress with minimal reliance on labeled data. Evaluation on the real-world Lakhimpur Kheri dataset demonstrates high accuracy 97.98\% for water stress, 85.08\% for nitrogen stress, and 83.47\% for combined water–nitrogen stress highlighting the robustness of the proposed approach in real agricultural settings. These results emphasize the strength of using lightweight self-supervised models for scalable and generalizable crop stress monitoring. 

Future work will aim to generalize the framework across diverse crop types and geographic regions with heterogeneous stress patterns. Integration of multimodal sensing data from UAVs, SAR sensors, and thermal imagery could further enrich feature representations and improve stress interpretability.
Exploring temporal attention mechanisms or transformer-based backbones may also yield better stress pattern recognition across longer crop cycles. Finally, deployment of the model in a real-time decision-support pipeline for farmers and agronomists remains a promising direction for translational impact.
\ifCLASSOPTIONcaptionsoff
  \newpage
\fi



%
\bibliographystyle{IEEEtran}
\bibliography{bibliography}

%








\end{document}